# Optimal Predefined-time Trajectory Planning for a Space Robot

Wen Yan[1] and Yicheng Liu[1]

[1] College of Electric Engineering, Sichuan Univesity, Chengdu 610065, China.

## Abstract

With the development of human space exploration, the space environment is gradually filled with abandoned satellite debris and unknown micrometeorites, which will seriously affect capture motion of space robot. Hence, a novel fast collision-avoidance trajectory planning strategy for a dual-arm free-floating space robot (FFSR) with predefined-time pose feedback will be mainly studied to achieve micron-level tracking accuracy of end-effector in this paper. However, similar to control, the exponential feedback results in larger initial joint angular velocity relative to proportional feedback. Therefore, a GA-based optimization algorithm is used to reduce the control input, which is just the joint angular velocity. Firstly, a pose-error-based kinematic model of the FFSR will be derived from a control perspective. Then, a cumulative dangerous field (CDF) collision-avoidance algorithm is applied in predefined-time trajectory planning to achieve micron-level collision-avoidance trajectory tracking precision. In the end, a GA-based optimization algorithm is used to optimize the predefined-time parameter to obtain a motion trajectory of low joint angular velocity of robotic arms. The simulation results verify our conjecture and conclusion.

## Introduction

In order to keep pose of base stable, a motion planning strategy for the balance arm is adopted in the trajectory planning algorithm to offset dynamic coupling by some researchers. Agrawal and Shirumalla [10] proposed an iterative-search-based dual-arm motion planning method to stabilize the position of base, but attitude was not considered. Yan et al. [12] studied the concept of base centroid virtual manipulator to reduce the pose error of base further. However, the above-mentioned methods cannot eliminate the errors of base and end-effector to zero theoretically. Moreover, when kinematic singularity occurs or the obstacle needs to be avoided, the error caused by additional algorithms will be further increased, and the previous method will be of little use. Fortunately, Liu et al. [13,14] presented a trajectory planning strategy with pose-feedback to eliminate the errors caused by singularity-avoidance and self-collision-avoidance method, but it cannot deal well with large cumulative error and the convergence time can also not be predicted.

Similar to control, because the essence of the feedback-based trajectory planning is to regard joint angular velocity of robotic arm as the control input and the velocity-level pose error obtained from the pose-error-based kinematic equation as the system output, the increase of joint angular velocity will be inevitable if exponential feedback is introduced to replace proportional feedback. Yoshio Yokose [15] applied genetic algorithm in the trajectory planning to obtain smooth joint angular velocity trajectory, but the tracking error is not considered primarily. Chen et al. [16] designed a considerate objective function of GA, but the tracking error will be large if there is a singular configuration in planned trajectory. Wu et al. [17] presented a GA-based singularity-free path planning for a space robot, and it can optimize the additional algorithm. However, none of the above methods can solve the problem of error elimination theoretically, and the error will inevitably increase if there are external interference, such as singularity and obstacle. Therefore, we try to solve the error problem theoretically by exponential feedback to obtain the high-precision motion trajectory of end-effector, and GA optimization strategy is only used to optimize the joint angular velocity trajectory to get the best trajectory planning strategy for the space robot.

Motivated the above issues, a GA-based optimal predefined-time trajectory planning for a FFSR is studied in this paper. The main purpose of this paper is to study how to obtain an optimal trajectory planning scheme for a space robot, which it is a continuing study. Therefore, the space robot model will adopt the more mature pose-error-based kinematic model that we studied earlier [13]. The main contributions of this paper lie in two aspects:

1) A novel GA-based predefined-time stability system method is studied to get the optimal parameter, which makes joint angular velocity smaller.

2) A predefined-time trajectory planning method for a FFSR system is proposed to stabilize the pose of base and achieve micron-level tracking precision in predefined time with dynamic coupling, singularity and obstacle.

In the following, Section 2 notes the related theoretic preliminaries. Section 3 introduces the pose-error-based kinematic model of a free-floating dual-arm space robotic system. Section 4 shows trajectory planning for the robotic system. Section 5 shows the simulation and analysis of the recent methods and the proposed method. Finally, section 6 presents the conclusion.

## Preliminaries

**Definition 1** [18]**:** Consider a system:

$$\dot{\mathbf{x}}(t) = f(\mathbf{x}(t)), \mathbf{x}(0) = \mathbf{x}_0. \tag{1}$$

where the state variables of system are set to $\mathbf{x} = [x_1, x_2 \cdots x_n]^T \in \mathbb{R}^n$, $f(\mathbf{x}): D \to \mathbb{R}^n$ is a continuous function, and $f(0) = 0$. If the above system is finite-time stable and the minimum upper bound of the settling time can be predicted to be $T_c$, the equilibrium in system (1) will be said to be predefined-time stable.

**Lemma 1** [18]**:** Consider the following system:

$$\dot{\mathbf{x}} = -\frac{1}{mT_c} \exp(\|\mathbf{x}\|^m) \frac{\mathbf{x}}{\|\mathbf{x}\|} \tag{2}$$

where $0 < m < 1$, and $T_c$ is a positive time parameter. The system in (2) is predefined-time stable and the convergence time maximum can be predefined as $T_c$ by Definition 1.

**Remark 1:** A predefined-time function can be defined as:

$$\boldsymbol{\varphi}(\mathbf{x}; T_c, m) = \frac{1}{mT_c} \exp(\|\mathbf{x}\|^m) \frac{\mathbf{x}}{\|\mathbf{x}\|} \tag{3}$$

## Trajectory Planning

A redefined-time stability system in Lemma 1 is applied in the time derivation of pose error of end-effector to obtain an exponential convergence, which can be expressed:

$$\dot{\mathbf{e}}_1 = -\boldsymbol{\varphi}(\mathbf{e}_1; T_c, m), \tag{9}$$

$$\dot{\mathbf{e}}_0 = -\boldsymbol{\varphi}(\mathbf{e}_0; T_c, m). \tag{10}$$

Then, according to the equation (4), (5), (9) and (10), the joint angular velocity trajectory of the FFSR with predefined-time convergence can be deduced:

$$\dot{\Theta}^1 = (J_m^1)^{-1}\big(J_e^{-1}(J_{ed}V_{ed} - \varphi(e_1; T_c, m)) - J_0 V_0\big), \quad (11)$$

$$\dot{\Theta}^2 = (J_c^2)^{-1}(J_c^1 \dot{\Theta}^1 - \varphi(e_0; T_c, m)J_b^{-1}H_0 - C). \quad (12)$$

A predefined-time stability is also applied in the pose-feedback-based-kinematic equations (16) and (17) to get the following collision-avoidance joint angular velocity trajectory of Arm-j (j=1,2):

$$\dot{\Theta}^1 = \mu \cdot (J_m^1)^T \cdot \text{diag}\left(\sum_{i=1}^{6} \frac{\sigma_i^1}{\sigma_i^1 + (\lambda_i^1)^2} v_i^1 (u_i^1)^T\right)(J_e^{-1}(J_{ed}V_{ed} - \varphi(e_1; T_c, m))$$

$$-J_0 V_0) + [E - \mu(J_m^1)^T(J_m^1(J_m^1)^T)^{-1}J_m^1]\dot{\Theta}_c^1 \quad (19)$$

$$\dot{\Theta}^2 = \mu \cdot (J_c^2)^T \cdot \text{diag}\left(\sum_{i=1}^{6} \frac{\sigma_i^2}{\sigma_i^2 + (\lambda_i^2)^2} v_i^2 (u_i^2)^T\right)(J_c^1 \dot{\Theta}^1 - \varphi(e_0; T_c, m)J_b^{-1}H_0 - C)$$

$$+[E - \mu(J_c^2)^T(J_c^2(J_c^2)^T)^{-1}J_c^2]\dot{\Theta}_c^2 \quad (20)$$

**GA-based Optimal Trajectory Planning for a Dual-arm FFSR**

Even if we adopt obstacle-avoidance and singularity-avoidance strategies in trajectory planning for the space robot, there is still a big problem that restricts the mission success of capture, which is the over-saturation of the robotic joint angular velocity. The maximum joint angular velocity of the mechanical system has to be controlled within a limited range, and its durability will be seriously affected if the robotic arm stays at a large joint angular velocity for a long time. Even the gear may be twisted to break in that case. Hence, an optimization of joint angular velocity should be adopted in the dual-arm FFSR system to get a smaller and smoother joint angular velocity of robotic arm. In this study, a GA-based optimal strategy is adopted in trajectory planning process, and the objective function **B** can be established as the following:

$$\mathbf{B} = \gamma(\alpha \int \Psi(\|\dot{\Theta}^1\|, \dot{\Theta}_{danger}^1) + \beta \int \Psi(\|\dot{\Theta}^2\|, \dot{\Theta}_{danger}^2)) + \Theta_b \quad (21)$$

where $\alpha$, $\beta$ and $\gamma$ are the weight parameters, $\dot{\Theta}_{danger}^j$ represents the an estimate that the joint angular velocity of Arm-j is approaching saturation. if $\mathbf{A} > \mathbf{B}$, $\mathbf{A} = \Psi(\mathbf{A}, \mathbf{B})$, else $0 = \Psi(A,B)$. The limiting factor of joint angular velocity $\Theta_b$ can be given as:

$$\Theta_b = \begin{cases} +\infty, & \dot{\Theta}^2 > \dot{\Theta}_{+max} \\ +\infty, & \dot{\Theta}^2 < \dot{\Theta}_{-max} \\ 0, & \dot{\Theta}_{-max} < \dot{\Theta}^2 < \dot{\Theta}_{+max} \end{cases} \quad (22)$$

in which $\dot{\Theta}_{\pm max}$ is the value of over-saturation of joint angular velocity $\dot{\Theta}^j$.

The genetic optimal algorithm can be described as:

**Step a** (the initialization of genetic population):

The population is composed by the two-dimensional array of predefined-time parameter m and $T_c$, and the size of population can be denoted by $S_{max}$. Then, $S_{max}$ joint angular velocity trajectories of dual-arm will generate to form the initial population. In the end, the joint angular velocity trajectories in initial population will be substituted into the fitness function, which is just the objection function **B**, to obtain the initial best individual fitness.

**Step b** (evaluation):

The purpose of evolution is to imitate the genetic inheritance of the biological population in nature and get a new generation by selecting mating, crossing and genetic variation. The fitness function is then used to select the best individual of the next generation.

**Step c** (selection):

The selection of this GA algorithm is roulette selection strategy. The probability of fitness can be expressed as:

$$\mathbf{P}_{si} = \frac{\mathbf{B}(\mathbf{x_i})}{\sum_{l=1}^{S_{max}} \mathbf{B}(\mathbf{x_l})} \tag{23}$$

where $\mathbf{x_i}$ is the i-th individual, and $\mathbf{x} = [\dot{\Theta}^1 \quad \dot{\Theta}^2]$. It is clear that the individual of will high fitness probability will be selected more times, because the gambling ball will be more likely to swing into a larger roulette area.

**Step d** (cross):

After selection, two individual are selected to cross by the crossing probability $P_c$, but the cross-point of individuals are chosen randomly.

**Step e** (mutation):

After cross, two individual are selected to mutate by the mutating probability $P_m$, but the cross-point of individuals are chosen randomly.

**Step f** (new population):

After first generation of evaluation, we will continue to evolve in G generations until the curves for average fitness and best fitness converge.

## Simulation

The initial momentum of the FFSR system **C** are set to 0, the simulation time is 20s. The initial position and attitude of base are $\mathbf{p_{b0}} = [-0.2832 \quad 0.3107 \quad 0.3248]^T$(m) and $\boldsymbol{\sigma_{b0}} = [1 \quad 0 \quad 0 \quad 0]^T$, respectively. The initial position and attitude of end-effector can be set to: $\mathbf{p_{e0}} = [-0.2832 \quad 0.3107 \quad 0.3248]^T$(m) and $\boldsymbol{\sigma_{e0}} = [0 \quad 0.8191 \quad 0 \quad 0.5736]^T$, respectively.

The initial joint angle of Arm-j are $\boldsymbol{\Theta}_0^1 = [0 \quad 47.72 \quad -93.91 \quad 0 \quad -23.82 \quad 0]^T(°)$ and $\boldsymbol{\Theta}_0^2 = [0 \quad -47.72 \quad 176.09 \quad 0 \quad -23.82 \quad 0]^T(°)$, respectively.

The position and attitude of the target are: $\mathbf{p_t} = [0.7147 \quad 0.4150 \quad -0.1758]^T$(m) and $\boldsymbol{\sigma_t} = [0.0215 \quad 0.9027 \quad 0.1184 \quad 0.4132]^T$, respectively. The position of the obstacle point is $\mathbf{p_o} = [-0.0691 \quad 0.6037 \quad 0.4742]^T$(m), and the danger radius is set to 0.2 (m).

The initial parameter of singularity-avoidance algorithm can be set to:

$\lambda_m = 0.08$, $\varepsilon^j = 0.02$.

As shown in the Figure 3, a velocity-level path planning of end-effector is adopted to get a desired trapezoidal velocity trajectory of end-effector.

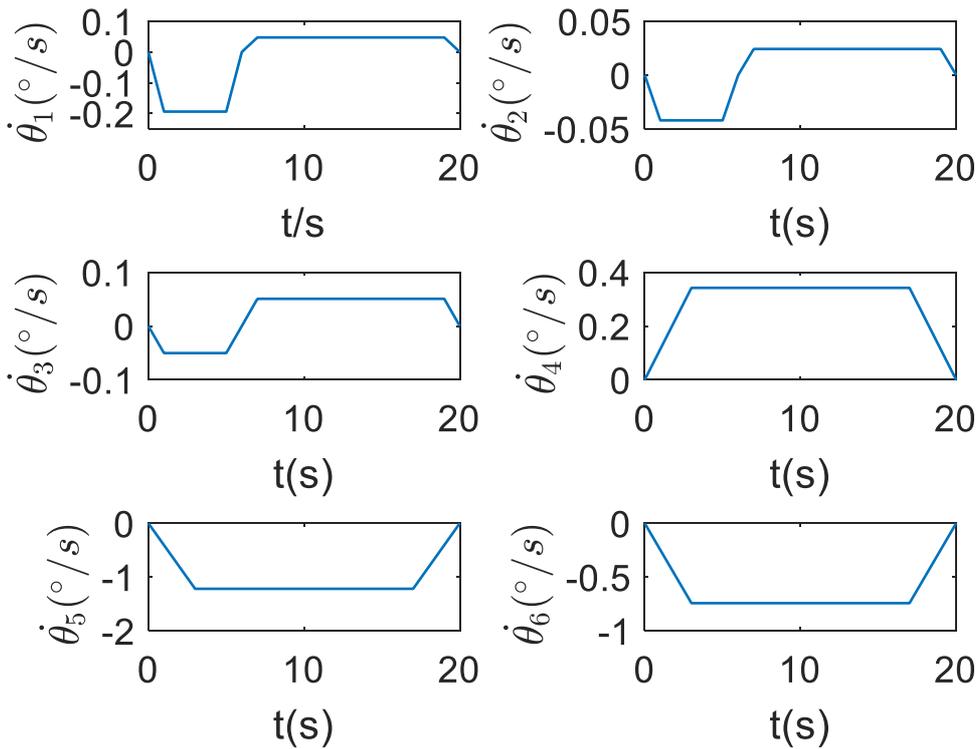

Figure 3. The desired trapezoidal velocity-level trajectory of end-effector

An optimal obstacle-avoidance trajectory planning method will be discussed in this simulation. The precision of the method should be high enough to improve the success rate of space-capture mission, and the time to return stability after avoiding obstacles should be predictable. Furthermore, similar to the finite time controller, the exponential convergence based pose feedback will also increase the joint angular velocity equivalent to control input. Therefore, we should optimize the parameters to achieve the minimum joint angular velocity trajectory.

**Pose-feedback-based Trajectory Planning for a Dual-arm FFSR with Predefined-time Convergence**

Different from singularity-avoidance algorithm [13], the accumulated error caused by obstacle avoidance is far greater than the instantaneous singularity, so the proportional-convergence-based pose feedback may not make this error converge to the micro-scale neighbourhood of zero, (or the convergence time is so long that the error cannot be eliminated). In order to compare the tracking accuracy of the proposed algorithm with the previous one, a trajectory planning method without obstacle-avoidance (no obstacle-avoidance algorithm), a no-pose-feedback obstacle-avoidance algorithm [20] (no pose feedback) and a CDF-based trajectory planning method with proportional-convergence-based pose feedback [14] (the proportional pose feedback) will be simulated to compare with the proposed method. The parameters of CDF-based obstacle-avoidance algorithm are set to: $\xi = 0.1$ and $\Delta = 17.25$. The predefined-time parameters are set to: $m = 0.1$ and $T_c = 3$.
5

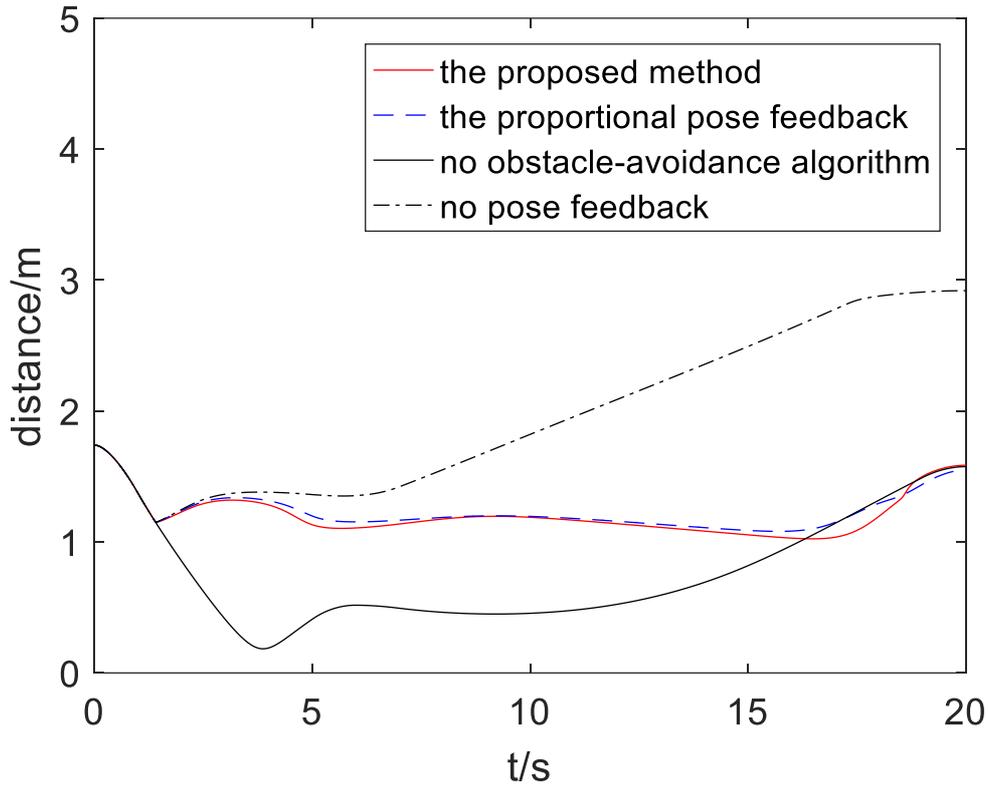

Figure 4. The curve of the closest distance between the obstacle point and the arm

As shown in the Figs. 4, the proposed method and other CDF-based trajectory planning method can keep the robotic arm from hitting the obstacle point, and the trajectory planning method without obstacle-avoidance algorithm will crash the robotic arm into an obstacle. According to the Fig. 5, it is clear that the proposed method and proportional pose-feedback-based method can make error converge to a small neighbourhood of zero, but the position error of end-effector in a trajectory planning method without obstacle-avoidance can only exist forever. Compared the proportional pose-feedback-based method with the proposed method, by Fig. 5, it is clear that the accumulated error caused by obstacle avoidance will be difficult to eliminate through proportional feedback, but exponential feedback is doing better. The final error of end-effector in proportional feedback is 0.033m, and the final error of end-effector in exponential feedback is $8.922 \times 10^{-12}$ m. Therefore, a predefined-time stability system should be applied in trajectory planning to enable the pose errors of end-effector to converge within a predetermined time $T_c$. The actual convergence time of position error of end-effector is only 1.198s, which is less than the estimated time $T_c = 3$s. Furthermore, as shown in the Fig 6, the attitude error of end-effector and the pose error of base can also be converged to the micron-scale neighbourhood of zero by the proposed algorithm, and the convergence time is also less than $T_c = 3$s.

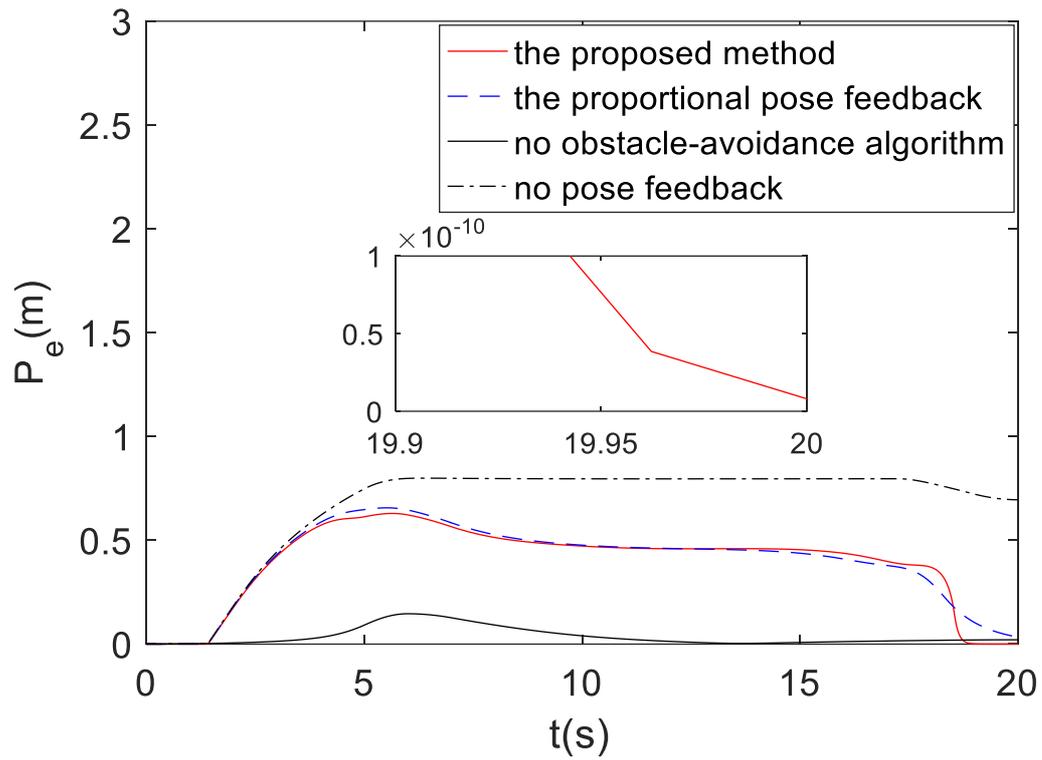

Figure 5. The position error curve of the end-effector (three axis synthesis)

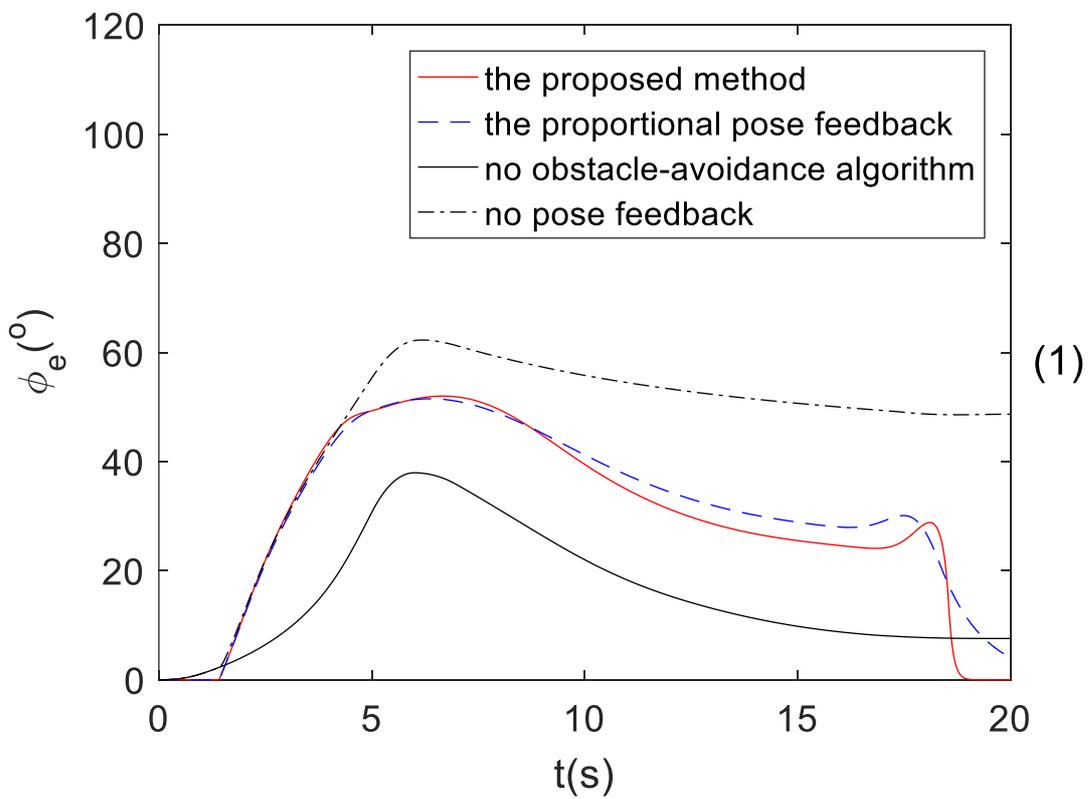

(1)

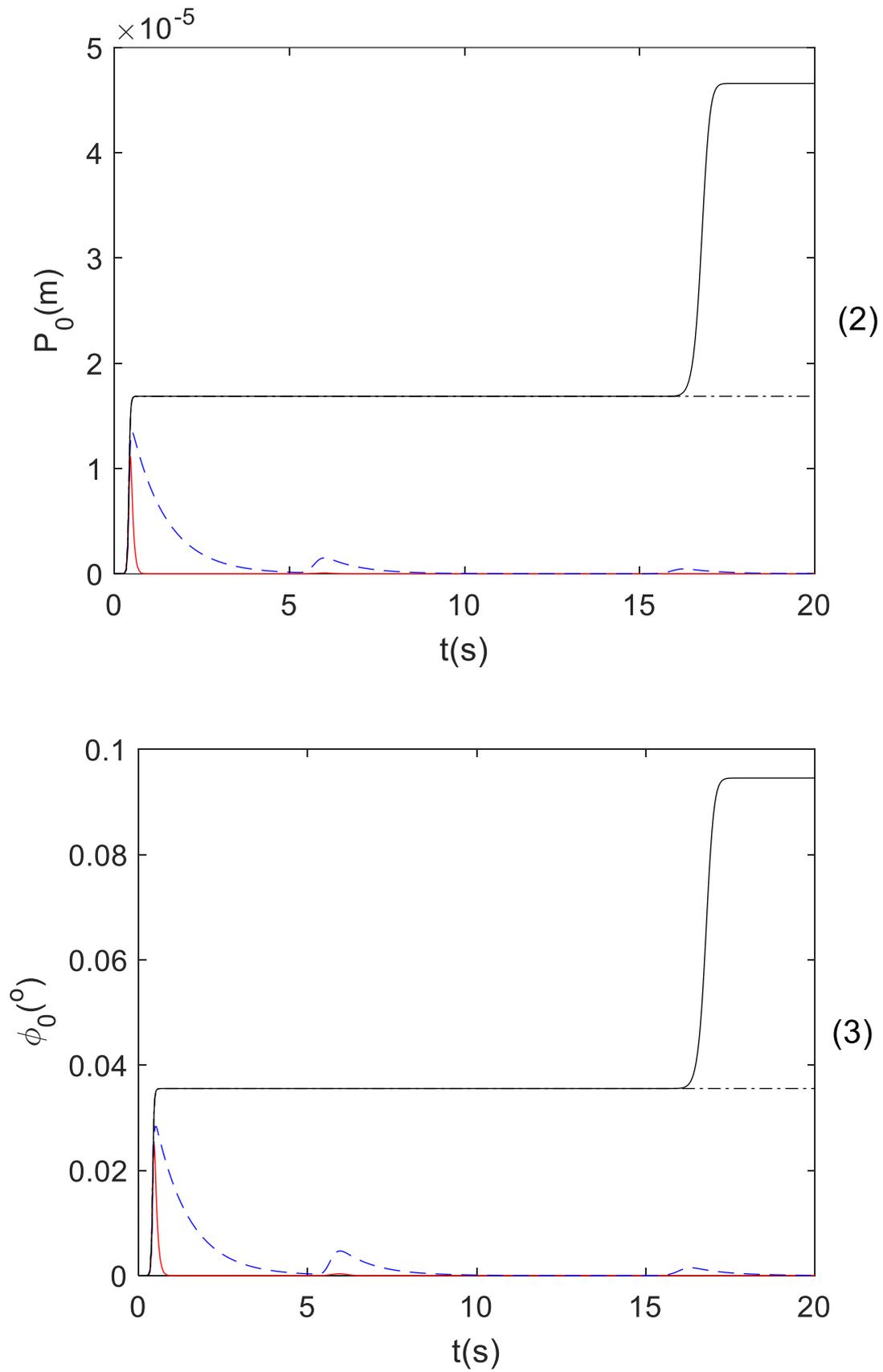

Figure 6. (1) The attitude error curve of the end-effector; (2) The position error curve of the base (three axis synthesis); (3) The position error curve of the base

# GA-based Optimal Micron-scale Predefined-time Trajectory Planning for a Dual-arm FFSR

Based on the above predefined-time trajectory planning method, we study a GA-based parameter optimization strategy. Similar to all finite-time feedback control, the terminal attractor in the exponential term will cause the initial control input, which is just the angular velocity of robotic arm, to become very large. Hence, in order to improve the durability of the manipulator, we need to optimize the parameters to get the best angular velocity trajectory of the joint.

The initial parameters of genetic algorithm can be set to:

$\alpha = 0.35$, $\beta = 0.65$, $\gamma = 0.0001$, $\dot{\Theta}^j_{danger} = 150°/s$ and $\dot{\Theta}_{\pm max} = \pm 200°/s$, $S_{max} = 100$, $P_c = 0.6$, $P_m = 0.1$, G=100.

The predefined-time parameter in no GA method can be set to:

$m = 0.1$, $T_c = 2.19$.

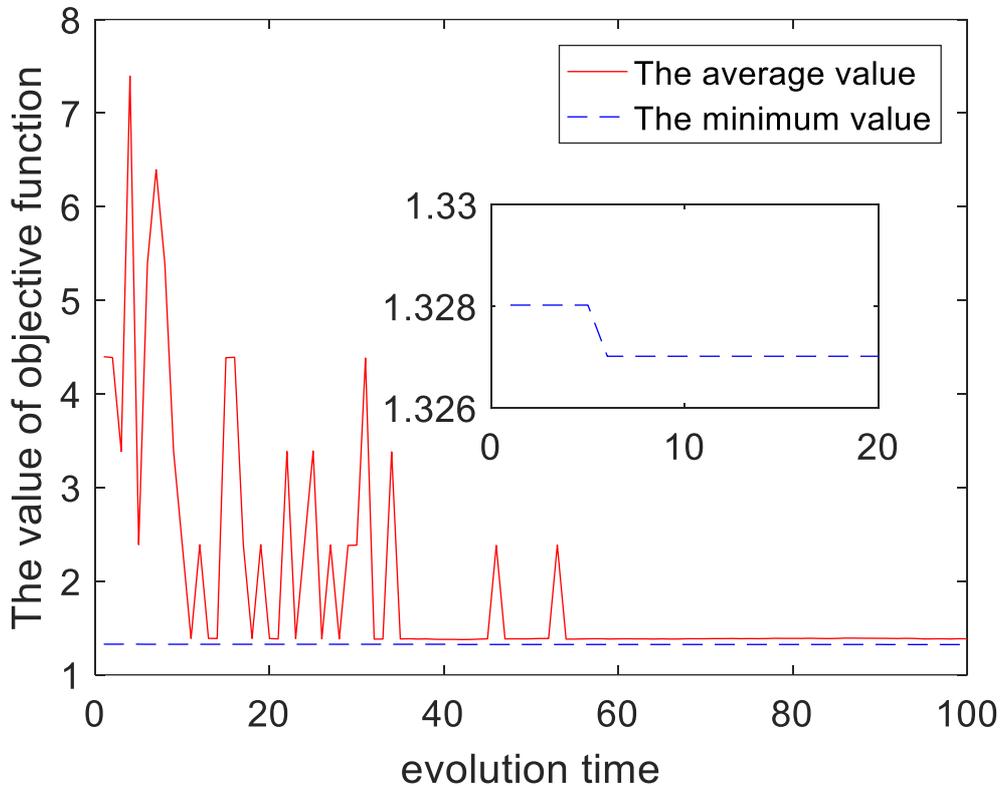

Figure 7. The each generation value of GA

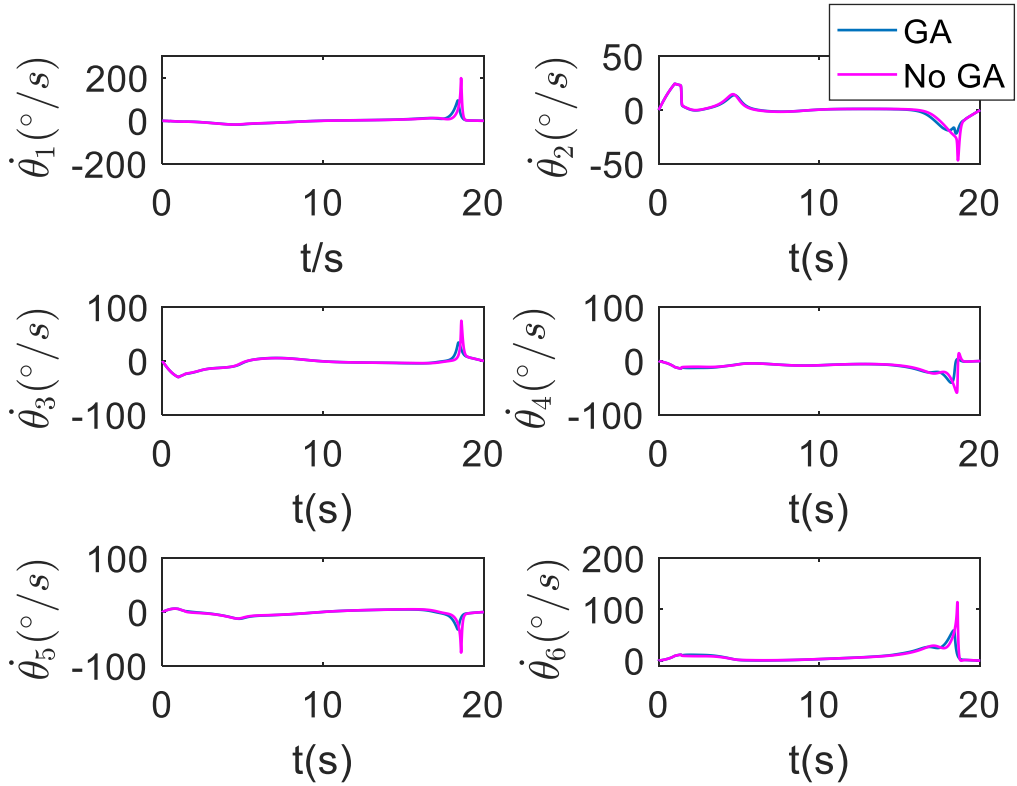

Figure 8. The joint angular velocity of mission arm

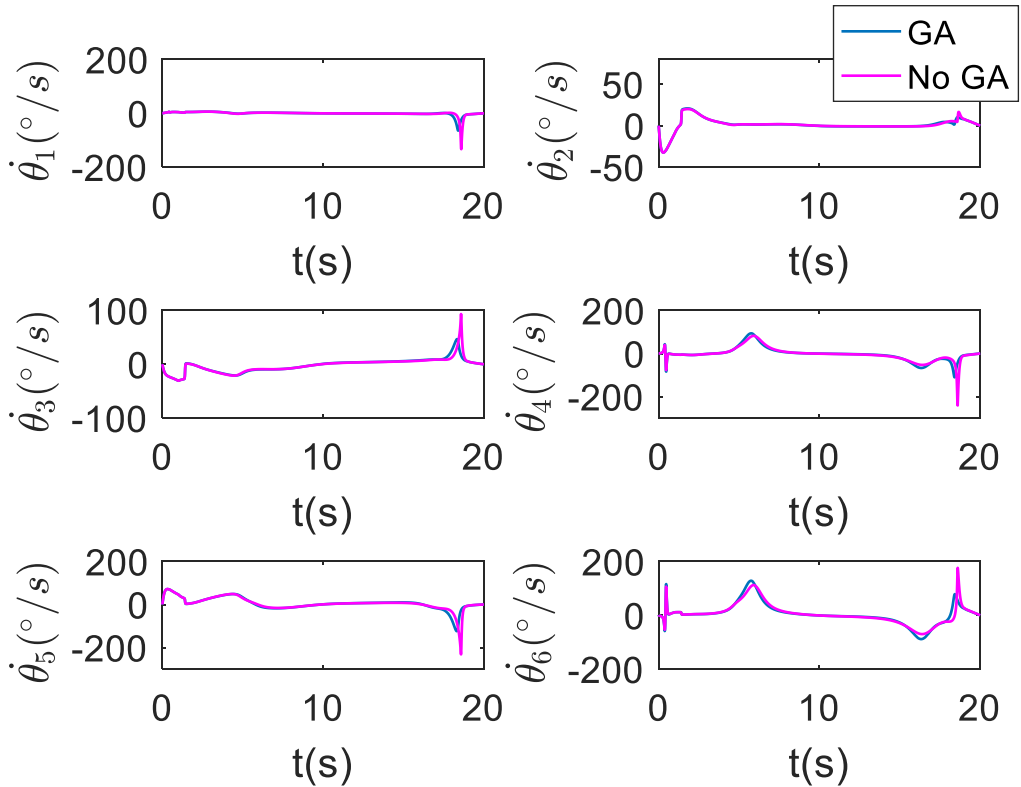

Figure 9. The joint angular velocity of balance arm

The comparison methods are the proposed GA-based algorithm and the predefined-time trajectory planning without optimal algorithm (for fair, the predefined-time function forms and time parameters $T_c$ of the comparison methods are consistent). According to the result in Fig. 7, the optimal predefined-time parameter can be set to $m = 0.2167$, $T_c = 2.19$. As shown in the Fig. 8 and 9, it is clear that the joint angular velocity in GA-based predefined-time trajectory planning method is smoother and smaller than the same method without optimal algorithm in whole. The max joint angular velocity of mission arm in GA-based method is $\theta_1^1 = 94.12°/s$, but the non-optimal method is $\theta_1^1 = 198.40°/s$. The max joint angular velocity of balance arm in GA-based method are $\theta_4^2 = -110.70°/s$ and $\theta_5^2 = -124.00°/s$, but the non-optimal method are $\theta_4^2 = -241.40°/s$ and $\theta_5^2 = -231.90°/s$. Furthermore, according to the Fig. 10, the micron-level pose tracking precision of end-effector can also be guaranteed. The final error of end-effector in exponential feedback is just $2.816 \times 10^{-7}$m, and the convergence time is 1.751s, which is less than the predefined time $T_c = 2.19$s.

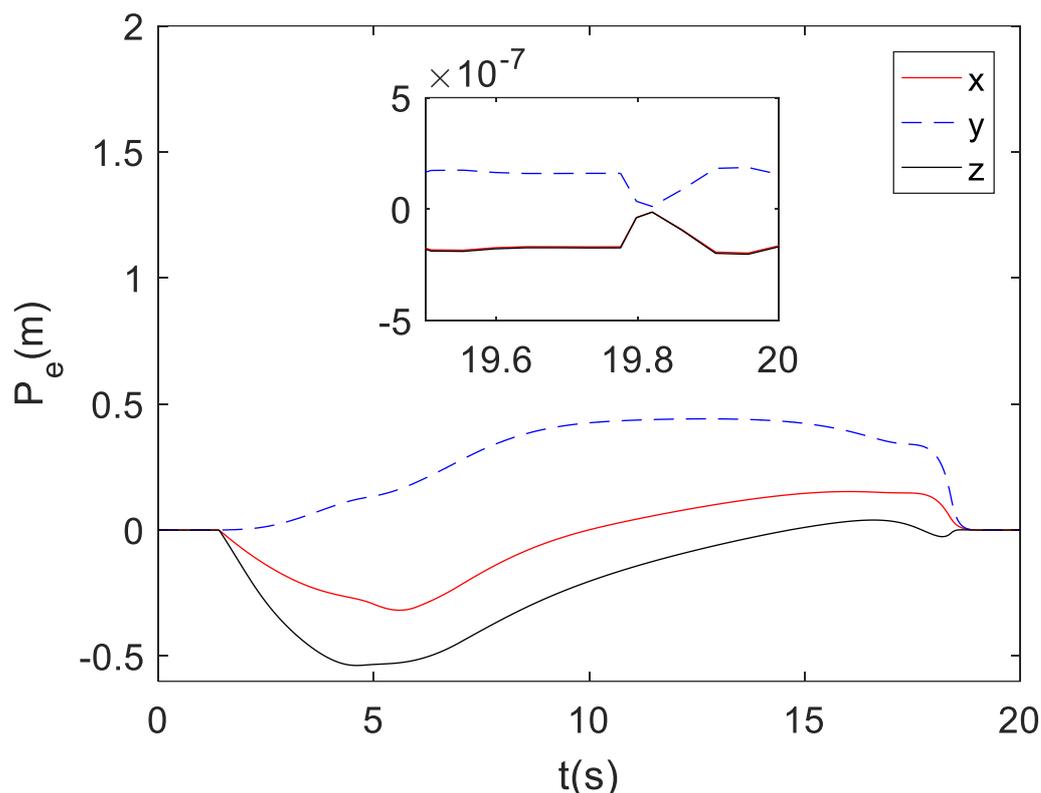

Figure 10. The position error curve of the end-effector in the proposed method with GA

## Conclusions

This paper addresses the optimization in predefined-time obstacle-avoidance trajectory planning for a space robot. In our previous study [13], the joint angular velocity trajectory planning of robotic arm was used to control the convergence rate of the pose tracking error of end-effector. However, if we apply exponential feedback in the pose-based kinematic equation to obtain an error rapid response form, the joint angular velocity of robotic arm, especially the balance arm, will become too large to damage the mechanism drive system. Hence, a GA-based optimal strategy should be adopted in trajectory planning for the space robot to obtain a small and smooth joint angular velocity trajectory. This work is different from the traditional optimization strategy of manipulator trajectory planning. We no longer consider the tracking error and smoothness of trajectory at the same time, but apply the predefined-time pose feedback to obtain the theoretical zero-error tracking precision so that only the smoothness of trajectory is considered. In the future, we will explore the machine-vision-based optimal trajectory tracking control for a space robot with a predefined-time-convergent neural network.

## Conflicts of Interest

The author claims that the copyright of the article research belongs to the authors, and other institutions without the authorization of both authors have no corresponding copyright.